\documentclass[sigconf]{acmart}
\usepackage[font=footnotesize]{caption}
\usepackage{threeparttable}
\setcopyright{none}
\usepackage{fancyhdr}
\usepackage[normalem]{ulem}
\usepackage{hyperref}
\usepackage{amsmath}
\usepackage{booktabs} 
\usepackage{upgreek}
\usepackage{graphicx}
\usepackage{verbatim}
\usepackage{multirow}
\usepackage{array}
\usepackage{subfigure}
\usepackage{float}
\usepackage{CJK}
\usepackage{setspace}
\usepackage{caption}
\usepackage{indentfirst}
\usepackage{dblfloatfix} 
\usepackage{multirow}
\usepackage[noend]{algpseudocode}
\usepackage{algorithmicx,algorithm}
\graphicspath{{./Figures/}}
\usepackage{tikz}
\usepackage{verbatim}

\newlength\szg     \newcommand\blackcir[1]{%
\settoheight\szg{#1}%
\tikz[baseline]{
\pgfmathparse{1}
\let\hfs\pgfmathresult
\filldraw (0,\szg/2) circle (\szg/2+0.30ex);
\node[white] at (0,\szg/2) {\makebox[0em][c]{\scalebox{\hfs}[0.8]{\textbf{#1}}}};
}}

\usepackage{fixltx2e}
\usepackage{hyperref}

\usepackage{nopageno}

\AtBeginDocument{%
  \providecommand\BibTeX{{%
    \normalfont B\kern-0.5em{\scshape i\kern-0.25em b}\kern-0.8em\TeX}}}
\begin{document}

\settopmatter{printacmref=false} 
\renewcommand\footnotetextcopyrightpermission[1]{} 
\pagestyle{plain} 


\title{\vspace{-1.3cm} Domain Knowledge-Infused Deep Learning for Automated Analog/Radio-Frequency Circuit Parameter Optimization}


\author{Weidong Cao$^{1,2}$,
   Mouhacine Benosman$^1$,
    Xuan Zhang$^2$,
    Rui Ma$^1$
    }
\affiliation{%
 \institution{$^{1}$Mitsubishi Electric Research Laboratories; $^{2}$Department of ESE, Washington University in St. Louis
     }
 }

\begin{abstract}
The design automation of analog circuits is a longstanding challenge.
This paper presents a reinforcement learning method enhanced by graph learning to automate the analog circuit parameter optimization at the pre-layout stage, i.e., finding device parameters to fulfill desired circuit specifications.
Unlike all prior methods, our approach is inspired by human experts who rely on domain knowledge of analog circuit design (e.g., circuit topology and couplings between circuit specifications) to tackle the problem.
By originally incorporating such key domain knowledge into policy training with a multimodal network, the method best learns the complex relations between circuit parameters and design targets, enabling optimal decisions in the optimization process.
Experimental results on exemplary circuits show it achieves human-level design accuracy ($\sim$\textbf{99}\%) with \textbf{1.5}$\times$ efficiency of existing best-performing methods.
Our method also shows better generalization ability to unseen specifications and optimality in circuit performance optimization.
Moreover, it applies to design radio-frequency circuits on emerging semiconductor technologies, breaking the limitations of prior learning methods in designing conventional analog circuits.

\end{abstract}

\maketitle

\section{Introduction}
\label{sec:intro}

Analog circuits play the fundamental role in processing analog signals and bridging the physical analog world and digital information world~\cite{cao1,cao2,cao3,cao4}.
Unlike digital circuits following standard automated design flows, analog circuit design relies on onerous human efforts and lacks effective design automation techniques at all stages.
Pre-layout design is one key stage in analog circuit design flow. 
It can be formulated as a parameter-to-specification (P2S) optimization problem, i.e., finding optimal device parameters (e.g., width and finger number of transistors) to meet desired circuit specifications (e.g., power and bandwidth) based on a pre-determined circuit topology.
This problem is very challenging as it seeks optimum parameters of diverse devices in a huge design space without exact rules.

Various automated techniques have been proposed for the P2S problem, mainly falling into optimization/learning-based category.
Optimization methods, e.g., Bayesian Optimization~\cite{bayesian} and Genetic Algorithm~\cite{genetic}, use corresponding algorithms to search for optimal device parameters.
They often suffer from several key issues, such as divergence, and re-starting from scratch if any change is made on given specifications.
Learning methods, i.e., supervised learning (SL) methods~\cite{supervised_2,op_amp_FCNN,cao5} and reinforcement learning (RL) methods~\cite{autockt_berke,GCN_RL_MIT}, have emerged recently.
They can achieve good convergence and cover a huge design space once well trained.
Despite the promise, these learning methods are still unable to reach human-level design accuracy, i.e., $\sim$100\%.
SL methods learn the static mapping between device parameters and circuit specifications~\cite{caophd}.
Due to the inherent approximation errors, they cannot ensure high design accuracy and endure weak generalization abilities with one-step inference.
RL methods learn a decision policy from state space of circuits to action space of device parameters and are often superior to SL methods via multi-step deployment.
However, without incorporating sufficient key state observations from environments into training, they fail to accurately learn the complex relations between device parameters and circuit specifications, leading to sub-optimal policies.
Moreover, existing learning methods cannot be applied to design more advanced analog circuits, e.g., radio-frequency (RF) circuits~\cite{CaoNN}, which require sophisticated time-consuming characterizations.
Without overcoming the issue, a much longer training time is needed by them before used for inference/deployment.

In this paper, we propose a domain knowledge-infused RL method to achieve human-level design accuracy and superior design efficiency for analog and RF circuits.
We are inspired by experienced human designers who leverage the key domain knowledge, e.g., topologies of circuits and couplings of specifications, to derive device parameters.
Particularly, they adopt a simplified circuit topology of a circuit, carefully consider design trade-offs between specifications, and use tens/hundreds of iterative fine tunings to seek the optimal circuit parameters.
Our RL method infuses the key domain knowledge into policy learning with a tailored multimodal policy network composed of a graph neural network (GNN) and a fully connected neural network (FCNN).
The GNN is built upon the topology of a given circuit.
It can capture the underlying physics of the circuit, e.g., device's connections and interactions.
The FCNN extracts the complex couplings of circuit specifications.
With such a unique policy network, our RL agent learns the best policy and makes optimal sequential decisions like a human expert to find device parameters.
Key contributions in the work are:
\begin{itemize}
  \vskip -12pt
	\item This paper presents the first domain knowledge-infused RL method to automate the P2S optimization of analog/RF circuit at the pre-layout level.
	\item This work proposes a unique multimodal policy network made of a circuit topology-based GNN and an FCNN to infuse key domain knowledge of circuit design into policy learning.
	\item The work also leverages transfer learning to notably accelerate RF circuits' design in a sophisticated and time-consuming simulation environment with the learned experiences from a coarse but time-efficient simulation environment.
	\item Experiments show the method achieves 99\% design accuracy, {1.5}$\times$ design efficiency of existing best-performing methods, a stronger generalization ability to unseen specifications, and better optimality in maximizing circuit's figure-of-merit.
\end{itemize}
\section{Background}
\label{sec:re_work}

\begin{figure}[!t]
\vskip -43pt
\includegraphics[width=0.75\columnwidth]{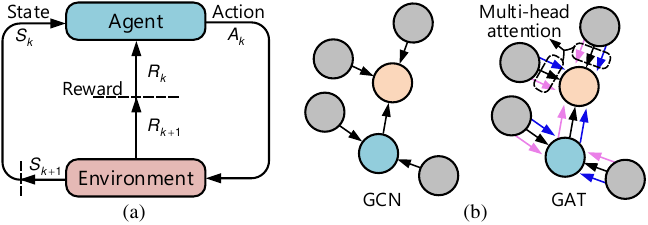}
\vskip -12pt
\caption{(a), A simplified illustration of RL. (b), A simplified illustration of a graph. Solid circles denote nodes and lines between circles represent edges.  The graph is processed by two GNNs: GCN and GAT.}
\label{fig: rl_intro}
\vskip -15pt
\end{figure}

\noindent{\textbf{Reinforcement Learning (RL):}} As shown in Figure~\ref{fig: rl_intro}(a), RL is an area of machine learning related to how an intelligent agent takes actions to maximize the cumulative return based on observed states from an environment .
In each episode, the agent starts from an initial state.
It then observes the state $S_k$ and takes an action $A_k$ based on a policy. 
Meanwhile, the environment updates a reward $R_{k+1}$ for that action and enters into a new state $S_{k+1}$.
The agent iterates through the episode with multiple steps, accumulating the reward at each step to obtain the final return.
With multiple episodes, the agent improves its decision quality and finally finds a well-learned policy to maximize the return.
The policy would be deployed for practical tasks, i.e., the agent follows the policy to finish a given task.
We apply RL to the P2S optimization of analog/RF circuits, which can best mimic the dynamic design process of human experts.

\noindent{\textbf{Learning with Graph Neural Networks (GNNs):}} GNNs~\cite{gan,gcn} 
directly learn the non-Euclidean data structure resembling a graph as shown in Figure~\ref{fig: rl_intro}(b). 
The graph is represented as $G=(V,E)$ with $V$ the set of node and $E$ the set of edge between connected nodes.
Assuming each node $v_i\in V$ has an $m$-dimensional vector of features, all node features form an matrix $X\in \mathbb{R}^{n\times m}$, $n=|V|$.
A GNN takes in $X$ as inputs and uses the class of each node in a graph or the class of an entire graph as labels.
Graph convolutional network (GCN)~\cite{gcn} and graph attention network (GAT)~\cite{gan} are two representative GNNs.
Compared to GCN, GAT has a multi-head attention mechanism on nodes as indicated in Figure~\ref{fig: rl_intro}(b) and can better learn high-dimensional complex relations between nodes.

Circuit topology is a graph and can be processed by GNNs.
A prior RL method~\cite{GCN_RL_MIT} uses GCN to process a circuit topology but has two key issues.
First, only a partial circuit topology is adopted by excluding power supply and bias nodes which, however, are the indispensable parts of a circuit graph. 
Second, the GCN node features are all static technology information, such as threshold voltage and electron mobility.
Without including the essential dynamic (variable) device parameters into node features, it is hard to learn the relations between device parameters and circuit specifications.
There are also several SL methods applying GNN to physical design~\cite{GNN_n0,GNN_n1} and electro-magnetic simulation~\cite{GNN_distributed} of analog circuits.
In contrast, our work harnesses GCN/GAT as a key part of our RL policy network to capture the physics of a given circuit topology, e.g., device's parameters, connections, and interactions, at the pre-layout stage.
We show that a GAT with the multi-head attention can better model a circuit topology than a GCN.

\section{Approach}
\label{sec:RL}

\begin{figure}[!t]
\vskip -43pt
\includegraphics[width=0.72\linewidth]{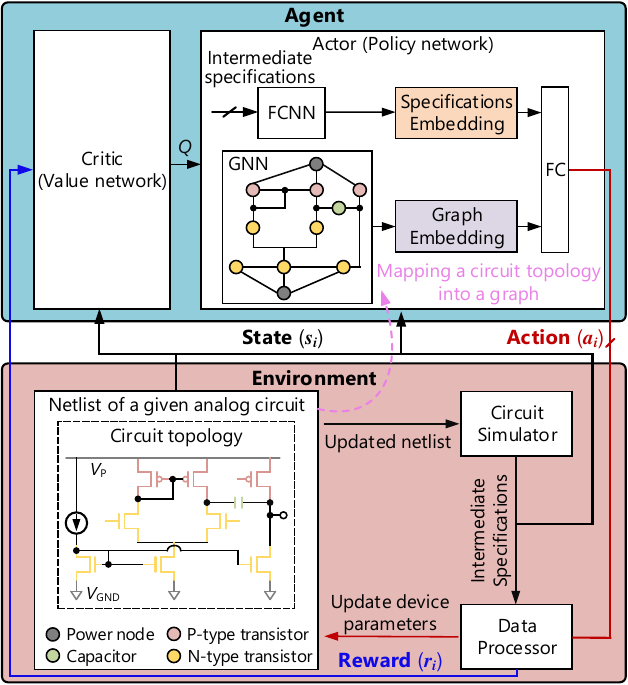}
\vskip -12pt
\caption{Overview of the RL method. The RL agent is based on an actor-critic method.
Our multimodal policy network consists of a circuit-topology-based GNN and an FCNN. We use a two-stage Op-Amp to show how to map a circuit topology into a graph.}
\label{fig: overview}
\vskip -21pt
\end{figure}

We target the P2S problem of analog/RF circuit design at the pre-layout stage and propose an RL approach for it.
Figure~\ref{fig: overview} shows the proposed RL method with the following five key elements.

\noindent{\textbf{Reward Function:}} The reward is directly related to the design goal.
We define the reward $r_i$ at each time step $i$ as 
\begin{equation}
 r_i= r, ~~\text{if} ~r < 0 ~~~ \text{or} ~~~  r_i=R, ~~\text{if} ~r =0,
\label{eq:reward}
\end{equation}
where $r=\sum\nolimits_{j=0}^{N-1}\min\{{(g^j_i-g^j_{*})}/{(g^j_i+g^j_{*})},0\}$ is a normalized difference between the intermediate specifications $g_i$ and the given specifications $g_{*}$.
The upper bound of $r$ is set to be 0 to avoid over-optimizing the parameters once the given specifications are reached.
All $N$ specifications are equally important.
We also give a large reward (i.e., $R=10$) to encourage the agent if the design goals are reached at some step.
The episode return $R_{s_0,g_{*}}$ of searching optimal device parameters for the given goals $g^{}_{*}$ starting from an initial state $s_0$, is the accumulated reward of all steps: $R_{s_0,g_{*}}= \sum\nolimits_{i=0} r_i$.
Our goal is to train a good policy to maximize $R_{s_0,g_{*}}$.

\noindent{\textbf{Action Representation:}} Inspired by human designers who iterate with fine-grained tuning steps to find optimal device parameters, we use discrete action space to  tune device parameters.
For each tunable parameter $x$ of a device (e.g., width and finger number of transistors), there are three possible actions at each step: increasing ($x+\triangle x$), keeping ($x+0$), or decreasing ($x-\triangle x$) the parameter, where ``$\triangle x$" is the smallest unit to update the parameter within its bound $[x_{\min},x_{\max}]$.
Assuming total $M$ device parameters, the output of the policy network is an $M\times 3$ probability distribution matrix with each row corresponding to a parameter.
The action is taken based on the probability distribution.

\noindent{\textbf{Environment:}}
A circuit design environment is used in this work.
It consists of a given circuit netlist, an industrial circuit simulator, such as Cadence Spectre or Keysight Advanced Design system (ADS) (for high-frequency RF circuits), 
and a data processing module (DPM).
As shown in Figure~\ref{fig: overview}, the simulator obtains intermediate circuit specifications at each time step.
The DPM then deals with the simulated results to feed back a reward to the agent using Eq.~\eqref{eq:reward}.
Meanwhile, it updates the device parameters to rewrite the circuit netlist based on the actions from the agent.

\noindent{\textbf{State Representation:}}
Capturing critical and adequate domain knowledge from the environment is key to training a good RL agent. 
In a circuit design environment, the circuit itself and the intermediate specifications are the main domain observations.
In our work, we for the first time adopt these two key practical observations to represent each state $s_i$.
We use a graph $G(V,E)$ to model the circuit based on its topology, where each node in set $V$ is a device and the connections between devices form the edge set $E$.
We also treat the power supply ($V_{\text{P}}$), ground ($V_{\text{GND}}$), and other DC bias voltages as extra nodes.
Figure~\ref{fig: overview} takes a two-stage operational amplifier (Op-Amp) as an example to show the mapping between its topology and the graph.
For a circuit with $n$ nodes, the state for the $k^{\text{th}}$ node is defined as its node feature $(t,\vec{p})$, where $t$ is the binary representation of the node type and $\vec{p}$ is the parameter vector of the node.
For transistors, the parameters are the width ($x_{\text{W}}$) and the finger number ($x_{\text{F}}$) while for capacitors, resistors, and inductors, the parameter is the scalar value of each device.
The parameter for power supply (ground or DC bias) is a voltage of $V_{\text{P}}$ (0 for $V_{\text{GND}}$ or $V_{\text{bias,}k}$ for bias node $k$).
Zero padding is used to ensure that the length of $\vec{p}$ for each node is the same. 
For a circuit with five different types of devices, two power nodes, one bias, the state of an N-type transistor is $[{0,0,1}, {x_{\text{W}},x_{\text{F}}}]$.
We also create a vector to represent intermediate specifications. 
For example, to design the Op-Amp, the state vector of specifications is expressed as $[G, B,PM,P]$ which are 
\text{gain} ($G$), \text{bandwidth} ($B$), \text{phase margin} ($PM$), and \text{power consumption} ($P$).

\noindent{\textbf{Agent:}} 
To incorporate the key domain knowledge into agent training such that it can make human-level decisions,
we propose a novel multimodal policy network for the agent based on actor-critic method~\cite{actor_critic} as shown in Figure~\ref{fig: overview}.
It consists of a circuit topology-based GNN and a fully connected neural network (FCNN), which is termed GNN-FC-based policy network.
The GNN is to distill the underlying physics (e.g., device’s types, parameters, and interactions) of a circuit graph into low-dimensional vector embedding.
While the FCNN takes the design goals as inputs to extract their coupled relations, i.e., design trade-offs.
The graph embedding and the FCNN embedding are then concatenated for further processing by the final fully-connected (FC) layers to update the actions.

We use GCN~\cite{gcn} and GAT~\cite{gan} to learn the embedding of circuit-level physical features respectively from the circuit graph $G=(V,E)$.
As an example, we show how to build the GCN below.
GAT~\cite{gan} can also be built similarly which is not elaborated here.
The node features of the $({l+1})^{\text{th}}$ layer in the GCN are obtained as 
 \vskip -6pt
\begin{equation}
H^{l+1}=f(H^l,A^{*})=\sigma(A^{*}H^lW^{l}).
\end{equation}
Here, $H^{l}\in \mathbb{R}^{n\times m_l}$ is the node feature matrix of the $l^{\text{th}}$ layer ($n$: number of nodes, $m_l$: feature dimension per node in the layer).
$H^{0}=X$ is the initial input node feature matrix.
$W^l$ is a weight matrix which combines the aggregated node features and pass them into a learnable layer (i.e, the $l^{\text{th}}$ layer) with a non-linear activation function $\sigma$ (i.e., $\tanh$ in our work).
$A^{*}$ is the matrix used to aggregate the neighbourhood features for a node, which is defined as: $A^{*}={\hat{D}}^{-{1}/{2}}\hat{A}{\hat{D}}^{-{1}/{2}},~\hat{A}=A+I$.
Here, $A$ is the adjacent matrix of the circuit graph; $I$ is an identity matrix; ${\hat{D}}$ is the diagonal node degree matrix of $\hat{A}$.
Using $A^{*}$ for aggregation is straightforward, as a device in a circuit graph is directly affected by its neighbors.
By stacking multiple GCN layers, one device can receive
information from farther devices that do not have a direct connection with it.

Combining the GNN, FCNN, and FC forms our policy network $\pi_{\theta}(a|s)$ parameterized by $\theta=\{W_{\text{GNN}}, W_{\text{FCNN}}, W_{\text{FC}}\}$ with $W_{\text{GNN}}$, $W_{\text{FCNN}}$, and $W_{\text{FC}}$ the learnable parameters of the GNN, FCNN and FC.
The value network preserves the same structure as the policy network except of the last layer.
It evaluates the actor's decision quality by yielding an estimation of the expected reward, $Q$, for the current policy execution.
The objective function of the problem can be formally defined as $J(\theta, G)={1}/{H} \cdot \sum\nolimits_{g\sim G}\mathbb{E}_{g,s\sim \pi_{\theta}}[R_{s,g}]$.
Here, $H$ is the the space size of all desired specifications $G$ and $R_{s,g}$ is the episode reward.
Our goal is to make the RL agent gain rich circuit design experiences by interacting with the environment.
Given the cumulative reward for each episode, we use Proximal Policy Optimization (PPO)~\cite{ppo} to update the parameters of the policy network as shown in Algorithm~\ref{alg: ppo} with a clipped objective below:
    \vskip -12pt
\begin{equation}
 L^{\text{CLIP}}(\theta) = \hat{\mathbb{E}}_i[\min(b_i(\theta), \text{clip}(b_i(\theta), 1-\epsilon, 1+ \epsilon ))\hat{A}_i],
\label{eq:ppo}
\end{equation}
where $\hat{\mathbb{E}}_i$ represents the expected value at time step $i$; $b_i$ is the probability ratio of the new policy and the old policy, and $\hat{A}_i$ is the estimated advantage at time step $i$.

\begin{algorithm}[!t]
\caption{Proximal Policy Optimization (PPO) Optimization}
\begin{footnotesize}
\begin{algorithmic}[1]
\State Input: initial policy parameters $\theta_0$ and initial value function parameters $\phi_0$
\For{$k=0,1,2,\cdots$}
    \State Collect a set of trajectories/episodes $\mathcal{D}_k=\{\tau_i\}$ by running policy $\pi_k=\theta_k$ in the circuit design environment.
    \State Compute rewards $\hat{R}_t$ for the trajectories/episodes.
    \State Compute advantage estimates, $\hat{A}_t$ 
    based on the current value function $V_{\phi_k}$.
    \State Update the policy by maximizing the PPO-clip objective in Eq.~\eqref{eq:ppo}, 
    via stochastic gradient ascent with Adam~\cite{adam}.
    \State Fit value function by regression on mean-squared error:
    \begin{equation}
 \phi_{k+1}= \arg \min{1}/({|\mathcal{D}_k|T})\cdot\sum\nolimits_{\tau\in\mathcal{D}_k}\sum\nolimits_{t=0}^T(V_{\phi}(s_t)-\hat{R}_t)^2,\nonumber 
\label{eq:ppo_va}
\end{equation}
    via stochastic gradient ascent with Adam~\cite{adam}.
\EndFor
\State \textbf{end for}
\end{algorithmic}
\label{alg: ppo}
\end{footnotesize}
\end{algorithm}

\begin{table*}[!t]
\vskip -0.6in
\centering
\caption{Design space of device parameters and sampling space of desired specifications of two circuit benchmarks.}
\begin{footnotesize}
\vskip -0.15in
\begin{tabular}{c|cccc|cc}
\toprule
Circuit types                                                       & \multicolumn{4}{c|}{Two-stage Op-Amp} & \multicolumn{2}{c}{RF PA} \\ 
\hline
Implementation technology & \multicolumn{4}{c|}{45 nm CMOS}       & \multicolumn{2}{c}{150 nm GaN}           \\ \hline
\# of device parameters  & \multicolumn{4}{c|}{$2\cdot7 + 1 =15$}      & \multicolumn{2}{c}{$2\cdot7=14$}               \\ \hline
\begin{tabular}[c]{@{}c@{}}Parameter constraints\\ (Design space) \end{tabular} &
  \begin{tabular}[c]{@{}c@{}}Width ($\mu$m)\\ $[1,100]$\end{tabular} &
  \begin{tabular}[c]{@{}c@{}}\# of fingers\\ $[2, 32]$\end{tabular} &
  \multicolumn{2}{c|}{\begin{tabular}[c]{@{}c@{}}capacitance (\textit{p}F)\\ $[0.1, 10]$\end{tabular}} &
  \begin{tabular}[c]{@{}c@{}}Width ($\mu$m)\\ $[16,100]$\end{tabular} &
  \begin{tabular}[c]{@{}c@{}}\# of fingers\\ $1, 2,...,16$\end{tabular} \\ \hline
\begin{tabular}[c]{@{}c@{}}Desired specifications\\ (Sampling space)\end{tabular} &
  \begin{tabular}[c]{@{}c@{}}Gain ($G$)\\ $[300, 500]$\end{tabular} &
  \begin{tabular}[c]{@{}c@{}}Bandwidth ($B$)\\ $[10^6, 2.5\cdot 10^7]$ Hz\end{tabular} &
  \begin{tabular}[c]{@{}c@{}}Phase margin ($PM$)\\  $[55^{\circ}, 60^{\circ}]$\end{tabular} &
  \begin{tabular}[c]{@{}c@{}}Power consumption ($P$)\\ $[10^{-4}, 10^{-2}]$ W\end{tabular} &
  \begin{tabular}[c]{@{}c@{}}Power efficiency ($E$)\\ $[50\%, 60\%]$\end{tabular} &
  \begin{tabular}[c]{@{}c@{}}Output power ($P$) \\ $[2,3]$ W\end{tabular} \\   \bottomrule
\end{tabular}
\end{footnotesize}
\label{tab1}
\vskip -9pt
\end{table*}

\begin{figure*}[!t]
\includegraphics[width=1.0 \linewidth]{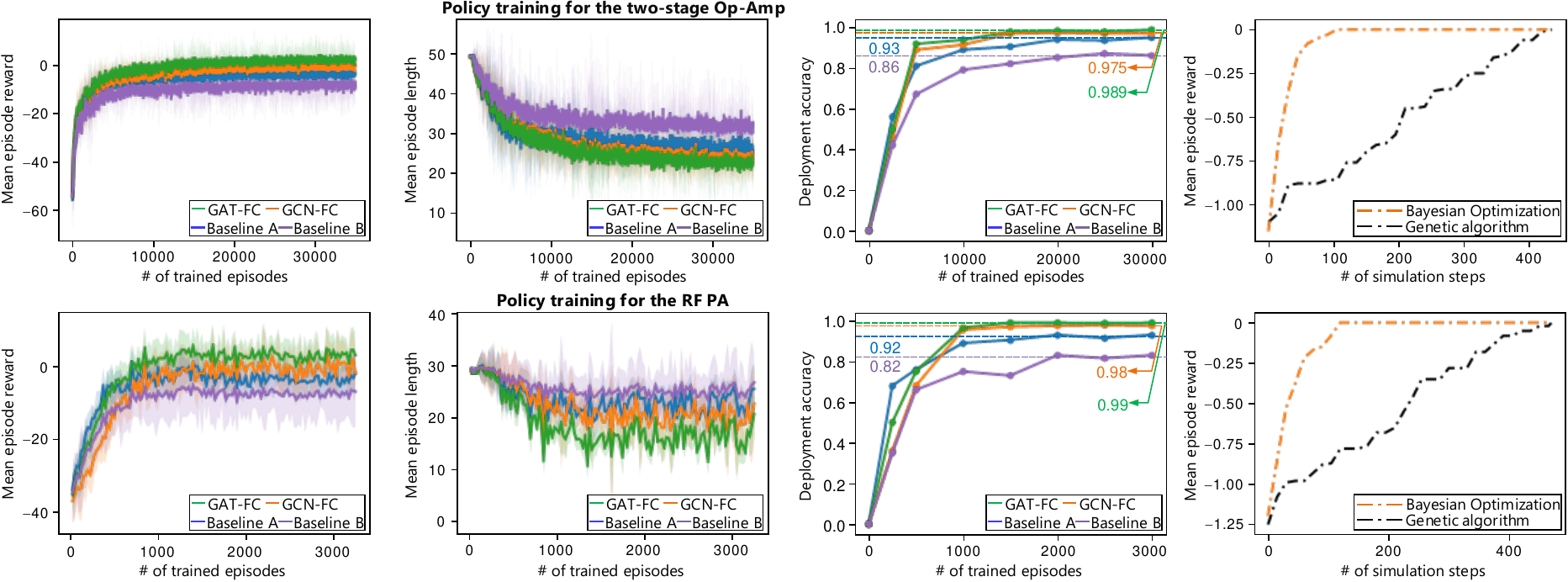}
\vskip -12pt
\caption{Comparing RL-based methods and optimization-based methods in the P2S problem with ours.
Two rows correspond to the two-stage Op-Amp and the RF PA. All results of RL methods are based on 6 random seeds.}
\label{fig: training_comp}
\vskip -12pt
\end{figure*}

\noindent{\textbf{Transfer Learning:}} We use transfer learning to speed up RF circuit design.
Generally, AC and DC simulations are sufficient to obtain all intermediate specifications $g_i$ at time step $i$ for low-frequency analog circuits (e.g., two-stage Op-Amps).
Such simulations are fast within tens of milliseconds in Cadence Spectre without delaying the learning of RL agents.
However, RF circuits (e.g., RF power amplifiers) often need more sophisticated  simulations to obtain accurate intermediate specifications which is timing-consuming.
Typically, one 
uses Harmonic Balance (HB) simulation ($\sim$1 minute/round in ADS) to attain intermediate specifications.
It significantly delays the reward calculation and training of RL agents.
To tackle the issue, fast ($\sim$1 second) but rough DC simulation is used to replace HB simulation.
It can obtain the not-very-accurate intermediate specifications for the quick approximation of the reward.
Our analyses show that the approximated rewards are often in $\pm$10\% error range compared to the ones obtained from the HB simulation.
Therefore, the learning process is remarkably speeded up. 
However, during the deployment stage for design automation, we still use HB simulation to guarantee the design quality and reliability.
In this way, the learned experiences from a coarse simulation environment can be accurately transferred into a fine simulation environment as verified by our results.
We think this may be due to the fact that a coarse design environment also provides sufficient information for the RL agent to learn the complicated relation between the device parameters and specifications.
For other advanced analog circuits, similar approximated rewards can also be obtained correspondingly.


\section{Experiments}
\label{sec:exp}

\begin{figure}[!t]
\begin{center}
\centerline{\includegraphics[width=0.8 \linewidth]{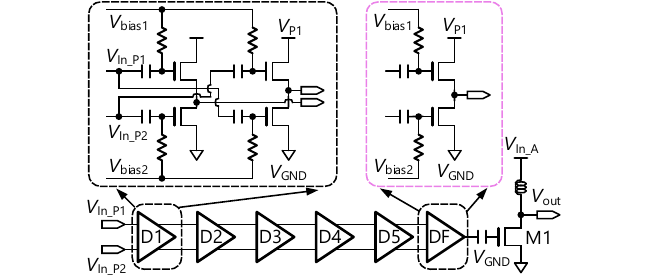}}
\vskip -12pt
\caption{Schematic of the RF PA~\cite{polar_trans} which consists of a driver stage (D1$\sim$D5 and DF) and a power amplifying transistor (M1).}
\label{fig: RF_PA}
\end{center}
\vskip -24pt
\end{figure}

Two circuits are used for evaluations.
One is the CMOS two-stage Op-Amp as shown in Figure~\ref{fig: overview}, which is a standard benchmark taken by prior methods~\cite{bayesian, autockt_berke, GCN_RL_MIT,genetic}. 
The other one is a gallium nitride (GaN) RF power amplifier (PA)~\cite{polar_trans} whose schematic is shown in Figure~\ref{fig: RF_PA}.
GaN is a promising alternative for conventional CMOS technology and for high-frequency power electronic applications~\cite{Gallium_nitride}.
The design space of device parameters and the sampling space of desired specifications for the two circuits are listed in Table~\ref{tab1}.

We adopt prior methods, i.e., Genetic Algorithm~\cite{genetic}, Bayesian Optimization~\cite{bayesian}, and RL methods~\cite{autockt_berke,GCN_RL_MIT} as our baselines.
These prior arts focus on two problems, i.e., P2S optimization~\cite{genetic, autockt_berke} and figure-of-merit (FoM) optimization~\cite{bayesian,GCN_RL_MIT}.
The prior RL methods exclude the key domain knowledge into policy learning and are not capable of designing RF circuits.
Baseline A (i.e., prior work~\cite{autockt_berke}) simply observes intermediate and given specifications from the environment and vectorizes them to train a feedforward policy network.
Baseline B (i.e., prior work~\cite{GCN_RL_MIT}) uses all static semiconductor technology information as observations to train a partial circuit topology-based GCN as the policy.
Such a method is often found to be divergent during training.
For conservative comparisons, we interpret and implement these RL arts~\cite{autockt_berke,GCN_RL_MIT} with our method.
First, we use the GCN design in our policy network as a more advanced implementation for the Baseline B.
Note that our GCN design is not only built upon a full circuit topology but also uses the essential dynamic (variable) device parameters as node features to better learn the relations between device parameters and circuit specifications.
Second, we build these RL baselines with the PPO technique~\cite{ppo} and discrete action space as done in our work as well as the transfer leaning technique to enable them to design RF circuits.
Our methods have two versions: GCN-FC policy and GAT-FC policy as GCN and GAT are respectively used as the GNN to capture the underlying physics of a full circuit topology.
We build circuit graphs using Deep Graph Library~\cite{dgl} and implement all methods with PyTorch.
We use equal amount of network parameters and the same set-ups for each baseline.
All our experiments are performed on an 8-core Intel CPU.
Moreover, supervised learning method~\cite{op_amp_FCNN} and a GAT-based implementation of Baseline B are also used as auxiliary comparisons with our method,
whose training results are not shown in the paper
but summarized in Table~\ref{tab2}. 



\noindent{\textbf{P2S Optimization:}}
Figure~\ref{fig: training_comp} shows the training curves (i.e., mean episode reward, mean episode length, and deployment accuracy) of
different RL methods for the P2S optimization.
The maximum episode length for each Op-Amp agent (RF PA agent) is set to be 50 (30). 
The total episodes used to train the two RL agents are chosen to be $3.5\cdot10^4$ and $3.5\cdot10^3$, respectively. 
As observed, our method achieves higher reward (first column), shorter mean episode length (second column), and higher deployment accuracy (third column) than all RL baselines.
Policy deployment applies a trained policy to automatically find the device parameters for given specifications.
Each point in Figure~\ref{fig: training_comp} (third column) comes from deploying each RL agent for 200 groups of randomly sampled specifications in Table~\ref{tab1}.

\begin{figure}[!t]
\vskip -42pt
\includegraphics[width=1.0 \linewidth]{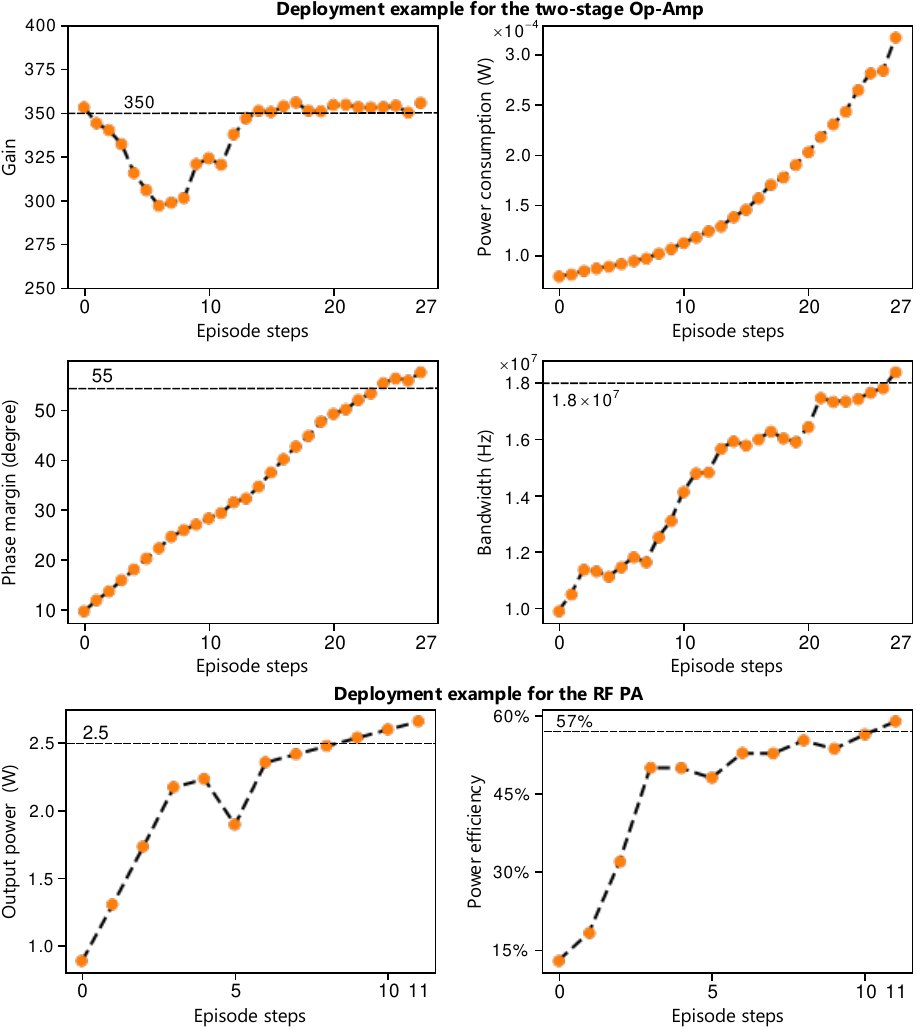}
\vskip -12pt
\caption{Deployment examples of the trained RL agent attempting to reach one group of the target specifications for each circuit.
}
\label{fig: amp_traj}
\vskip -15pt
\end{figure}


Given the desired circuit specifications, Genetic algorithm~\cite{genetic} and Bayesian Optimization~\cite{bayesian} use algorithms to guide the searching process by maximizing $r$ in Eq.~\eqref{eq:reward} without training.
The last column in Figure~\ref{fig: training_comp} shows the optimization curves. 
However, they cannot leverage transfer learning and have to use HB simulation to ensure design quality, which is time-consuming. 
We observe that Genetic Algorithm (Bayesian Optimization) often requires $\sim$400 ($\sim$100) steps/simulations to find optimal device parameters, incurring long run-time delay.
Moreover, due to the limitations, such as being stuck at a local optimum and even divergence, the algorithm cannot guarantee the correctness of each design.
Based on 30-group random experiments, the design accuracy is $76.7\%$ ($83.7\%$) for the Genetic Algorithm (Bayesian Optimization). 

The comparison shows that the our methods achieve the highest design efficiency (with fewer deployment steps per episode, $\sim$20 steps for Op-Amp and $\sim$15 steps for RF-PA) and human-level design accuracy (higher policy deployment accuracy, $99\%$) for both circuits design.
Particularly, we also note that the GAT-FC-based policy is superior to the GCN-FC-based policy.
Such a comparison shows that circuit topology is an important ingredient in RL-based policy learning.
And a better circuit topology modeling method, that is using GAT with the multi-head attention mechanism to learn higher-dimensional interactions among circuitry nodes, can further improve the performance of a policy.

\begin{table*}[!t]
\vskip -0.70in
\caption{Comparison summarization of different design automation methods.}
\begin{footnotesize}
\vskip -0.15in
\begin{threeparttable}
\begin{tabular}{c|c|c|c|c|c|c}
\toprule
\multirow{3}{*}{Methods} &
  \multicolumn{2}{c|}{\multirow{3}{*}{Sufficient key domain knowledge (?)}} &
  \multicolumn{3}{c|}{P2S optimization} &
  FoM optimization \\ \cline{4-7} 
 &
  \multicolumn{2}{c|}{} &
  \multirow{2}{*}{Design accuracy}&
  \multicolumn{2}{c|}{Mean \# of design steps} &
  FoM value \\ \cline{5-7} 
 &
  \multicolumn{2}{c|}{} &
   &
  Two-stage Op-Amp &
  RF PA &
  RF PA \\ \midrule
Genetic Algorithm~\cite{genetic} &
  \multicolumn{2}{c|}{NO} & 76.7\%
   & 370
   & 389
   & 2.53
   \\ \hline
Bayesian Optimization~\cite{bayesian} &
  \multicolumn{2}{c|}{NO} & 83.7\%
   & 86
   & 105
   & 2.61
   \\ \hline
Supervised learning~\cite{op_amp_FCNN} &
  \multicolumn{2}{c|}{NO} & 79\%
   & 1
   & 1
   & N/A
   \\ \hline
RL method (Baseline A)~\cite{autockt_berke} &
  \multicolumn{2}{c|}{\begin{tabular}[c]{@{}c@{}}NO
  \end{tabular}} & 92\%
   & 27
   & 23\tnote {a}
   & 2.92\tnote {a}
   \\ \hline
RL method (Baseline B)~\cite{GCN_RL_MIT} &  \multicolumn{2}{c|}{\begin{tabular}[c]{@{}c@{}}NO\tnote {b}\end{tabular}}
   & 84\% (87\%)
   & 32 (31)
   & 25 (23)\tnote {a}
   & 2.81 (2.86)\tnote {a}
   \\ \hline
\multirow{2}{*}{\textbf{Our RL method}} &
  \multirow{2}{*}{\begin{tabular}[c]{@{}c@{}} \textbf{YES: Full circuit topology}  + \\ \textbf{Specification couplings}\end{tabular}} & 
  \textbf{GCN + FCNN} & \textbf{98\%}
   & \textbf{24}
   & \textbf{19}
   & \textbf{3.18}
   \\ \cline{3-7} 
 &
   &
 \textbf{ GAT + FCNN} &\textbf{99\%}
   & \textbf{21}
   & \textbf{16}
   & \textbf{3.25}
   \\   \bottomrule
\end{tabular}
\begin{tablenotes}\scriptsize
\item[a] They originally cannot design RF circuits. We leverage our transferring learning technique to enable them to design RF circuits.
\item[b] Implemented with our GCN (GAT) part: full circuit topology $+$ device parameters as key node features.
\end{tablenotes}
\end{threeparttable}
\end{footnotesize}
\label{tab2}
\vskip -12pt
\end{table*}


\noindent{\textbf{Automated Design with Policy Deployment:}}
We take our GCN-FC-based policy as an example to show the deployment process.
Figure~\ref{fig: amp_traj} illustrates the deployment where RL agents automatically find optimal device parameters for a group of randomly sampled specifications (the horizontal dashed lines in each sub figure).
The sampled desired specifications for the two-stage Op-Amp are gain ($G=350$), bandwidth ($B=1.8\cdot10^7$ Hz), phase margin ($PM= 55^{\circ}$), and power consumption ($P=4\cdot 10^{-3}$ W). 
And for the RF PA, they are output power ($P=2.5$ W), and power efficiency ($E=57\% $).
Note that the smaller the power consumption is, the better the performance is.
At the initial state, the intermediate specifications ($y$-axis of each sub-figure) often deviate a lot from the desired ones. 
As the deployment continues, they get closer to the desired ones by following the trained policy.
An interesting phenomenon here is that when some specification is first achieved, the RL agent will not over-optimize it too much but instead try to optimize the remaining ones.
For example, the gain of the two-stage Op-Amp is first attained at the $14^{\text{th}}$ deployment step. 
In the following steps, the RL agent focuses on optimizing phase margin and bandwidth.
The similar analysis also applies to the design of the RF PA.
We also analyze a few failed cases where our trained policy cannot converge to the optimal device parameters.
We observe that in these cases, some specifications can converge to a neighborhood of the desired ones, but after which they deviate a bit from the goal.
Fortunately, we find that by slightly tuning the device parameters with manual effort at that particular step, the design goal is also easily achieved.
In this way, the design accuracy can be improved to 100\%.
These results show that human designers can still greatly benefit from the trained policy, if used as an efficient warm-start for the manual tuning, even if an automated deployment fails.

\begin{figure}[!t]
\vskip -42pt
\includegraphics[width=1.0 \linewidth]{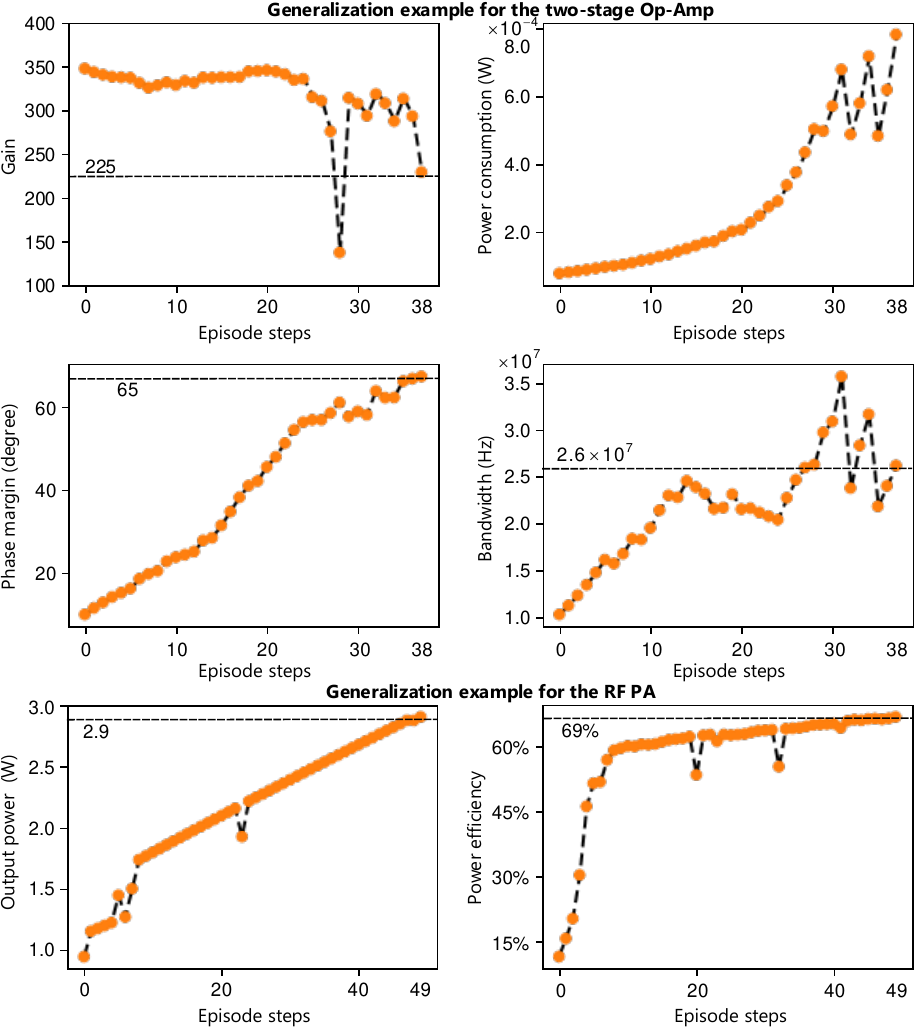}
\vskip -12pt
\caption{Generalization examples of the trained RL agent attempting to reach one group of the unseen new specifications for each circuit.
}
\label{fig: gene}
\vskip -15pt
\end{figure}


\noindent{\textbf{Generalization to Unseen Specifications:}}
We also evaluate the generalization ability of our GCN-FC-based policy by deploying it with unseen specifications out of the sampling space in Table~\ref{tab1}.
Figure~\ref{fig: gene} shows such an example, where the horizontal dashed lines denote these unseen specifications: gain ($G=225$), bandwidth ($B=2.6\cdot10^7$ Hz), phase margin ($PM= 65^{\circ}$), power consumption ($P=6\cdot 10^{-3}$ W) for the two-stage Op-Amp; output power ($P=2.9$ W), and power efficiency ($E=69\%$) for the RF PA.
Compared to policy deployment with the specifications coming from the sampling space, the deployment with unseen specifications usually requires more search steps.
For example, the generalization for the RF PA needs 49 steps to achieve the design goals while 11 steps are enough for the normal deployment in Figure~\ref{fig: amp_traj}. 
This is because that unseen specifications are beyond the scope of training datasets, thereby demanding more steps to reach optimal parameters.
We also analyze the generalization ability of baseline methods (not shown here) and find that they often do not generalize well as ours even with a higher number of search steps.
The better generalization ability of ours is attributed to the fact 
that it is capable 
of capturing key domain knowledge from state space, hence can better apply the learned experiences 
to unseen specifications
at the inference time.



\begin{figure}[!t]
\begin{center}
\centerline{\includegraphics[width=1.0 \linewidth]{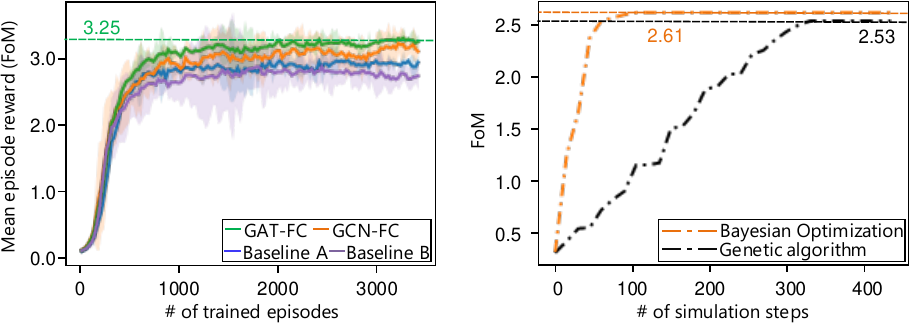}}
\vskip -12pt
\caption{Comparing FoM optimization between different methods. All results of RL methods are reported based on 6 random seeds.}
\label{fig: fom}
\end{center}
\vskip -21pt
\end{figure}

\noindent{\textbf{FoM Optimization:}}
We also compare
all methods in optimizing FoM 
by using the RF PA as an example.
To apply the methods to this problem, we use the FoM definition~\cite{bayesian} of RF PAs, i.e., $r_i=P_i+3\cdot E_i$ to revise the reward function in Eq.~\eqref{eq:reward}.
Here, $P_i,E_i$ are the intermediate specifications at time step $i$.
In the training, we use references $P_{r},E_{r}$ for normalization, i.e., $r_i={(P_i-P_{r})}/{(P_i+P_{r})}+3\cdot{(E_{i}-E_r)}/{(E_i+E_{r})}$.
For each RL method, we train the corresponding RL agent with $3.5\times 10^3$ episodes.
Figure~\ref{fig: fom} shows the optimization curves of all methods.
Our methods (GAT-FC/GCN-FC) obtains a higher FoM and the GAT-FC-based policy attains the highest one, showing the superiority of our methods.

\noindent{\textbf{Comparison Summarization:}}
We summarize the comparisons
in Table~\ref{tab2}.
In tackling the P2S optimization, our method achieves the highest design accuracy.
Optimization methods~\cite{genetic,bayesian} cannot ensure a high design accuracy because of their limitations, e.g., being stuck at a local optimum (caused by non convexity) or divergence of the algorithms. 
Due to the inherent approximation errors, SL methods~\cite{op_amp_FCNN} suffer from a low design accuracy.
RL methods~\cite{autockt_berke,GCN_RL_MIT} excluding the key domain knowledge cannot reach the human-level design accuracy as ours.
Due to such limitations, these methods show a weaker generalization ability than ours, either.
Despite not excelling the design efficiency of SL methods with one-step prediction, once trained our method uses fewer steps to find the optimal device parameters for the same desired specifications, improving the design efficiency by average $1.5\times$ compared to the prior RL methods~\cite{autockt_berke,GCN_RL_MIT} and average $10\times$ compared to optimization methods~\cite{genetic,bayesian}.
In the application of FoM optimization, our method also achieves higher FoM value than prior RL methods and optimization methods.
In summary, our RL method inspired by key domain knowledge of analog circuit design and human-like multiple tuning steps achieves the best balance between the design accuracy and efficiency as well as the best optimality.

\section{Conclusion}
We have shown a learning method for the automated design of analog circuits.
The key property of our framework is to incorporate domain knowledge of analog circuit design (i.e., the underlying physical topology of a given circuit and the trade-offs between specifications) into the proposed combined GNN (GCN/GAT)-FC-based multimodal policy network.
We show that such a method is superior to other methods without such considerations in designing various analog circuits with higher accuracy, efficiency, and optimality.
We expect that our method will assist IC industry to accelerate the analog/RF chip design, with artificial agents that master massive circuitry optimization experiences via continuous learning.

\noindent{\textbf{Acknowledgment:}} Weidong Cao was an intern at MERL. This work is supported by MERL with additional support for Weidong Cao, Xuan Zhang in part by National Science Foundation grant no. CCF-1942900.
This paper is accepted by 2022 Design Automation Conference (DAC).

\bibliographystyle{unsrt}
\bibliography{main}

\begin{thebibliography}{10}

\bibitem{cao1}
Weidong Cao, Xin He, Ayan Chakrabarti, and Xuan Zhang.
\newblock {NeuADC: Neural Network-Inspired Synthesizable Analog-to-Digital
  Conversion}.
\newblock {\em IEEE Transactions on Computer-Aided Design of Integrated
  Circuits and Systems}, 39(9):1841--1854, 2020.

\bibitem{cao2}
Weidong Cao, Xin He, Ayan Chakrabarti, and Xuan Zhang.
\newblock {NeuADC: Neural Network-Inspired RRAM-Based Synthesizable
  Analog-to-Digital Conversion with Reconfigurable Quantization Support}.
\newblock In {\em 2019 Design, Automation Test in Europe Conference Exhibition
  (DATE)}, pages 1477--1482, 2019.

\bibitem{cao3}
Weidong Cao, Liu Ke, Ayan Chakrabarti, and Xuan Zhang.
\newblock {Neural Network-Inspired Analog-to-Digital Conversion to Achieve
  Super-Resolution with Low-Precision RRAM Devices}.
\newblock In {\em 2019 IEEE/ACM International Conference on Computer-Aided
  Design (ICCAD)}, pages 1--7, 2019.

\bibitem{cao4}
Weidong Cao, Liu Ke, Ayan Chakrabarti, and Xuan Zhang.
\newblock {Evaluating Neural Network-Inspired Analog-to-Digital Conversion With
  Low-Precision RRAM}.
\newblock {\em IEEE Transactions on Computer-Aided Design of Integrated
  Circuits and Systems}, 40(5):808--821, 2021.

\bibitem{bayesian}
Wenlong Lyu, Fan Yang, Changhao Yan, Dian Zhou, and Xuan Zeng.
\newblock {Batch {B}ayesian Optimization via Multi-objective Acquisition
  Ensemble for Automated Analog Circuit Design}.
\newblock In {\em Proceedings of the 35th International Conference on Machine
  Learning}, pages 3306--3314, 2018.

\bibitem{genetic}
Bo~Liu, Yan Wang, Zhiping Yu, Leibo Liu, Miao Li, Zheng Wang, Jing Lu, and
  Francisco~V. Fernández.
\newblock {Analog Circuit Optimization System Based on Hybrid Evolutionary
  Algorithms}.
\newblock {\em Integration}, 42(2):137 -- 148, 2009.

\bibitem{supervised_2}
Y.~{Li}, Y.~{Wang}, Y.~{Li}, R.~{Zhou}, and Z.~{Lin}.
\newblock {An Artificial Neural Network Assisted Optimization System for Analog
  Design Space Exploration}.
\newblock {\em IEEE Transactions on Computer-Aided Design of Integrated
  Circuits and Systems}, 39(10):2640--2653, 2020.

\bibitem{op_amp_FCNN}
H.~{M.V.} and B.~P. {Harish}.
\newblock {Artificial Neural Network Model for Design Optimization of 2-stage
  Op-amp}.
\newblock In {\em 2020 24th International Symposium on VLSI Design and Test
  (VDAT)}, pages 1--5, 2020.

\bibitem{cao5}
Weidong Cao, Yilong Zhao, Adith Boloor, Yinhe Han, Xuan Zhang, and Li~Jiang.
\newblock {Neural-PIM: Efficient Processing-In-Memory with Neural Approximation
  of Peripherals}.
\newblock {\em IEEE Transactions on Computers}, pages 1--1, 2021.

\bibitem{autockt_berke}
K.~{Settaluri}, A.~{Haj-Ali}, Q.~{Huang}, K.~{Hakhamaneshi}, and B.~{Nikolic}.
\newblock {AutoCkt: Deep Reinforcement Learning of Analog Circuit Designs}.
\newblock In {\em 2020 Design, Automation \& Test in Europe Conference
  Exhibition (DATE)}, pages 490--495, 2020.

\bibitem{GCN_RL_MIT}
H.~{Wang}, K.~{Wang}, J.~{Yang}, and L.~{Shen} et~al.
\newblock {GCN-RL Circuit Designer: Transferable Transistor Sizing with Graph
  Neural Networks and Reinforcement Learning}.
\newblock In {\em 2020 57th ACM/IEEE Design Automation Conference (DAC)}, pages
  1--6, 2020.

\bibitem{caophd}
Weidong Cao.
\newblock Machine learning for analog/mixed-signal integrated circuit design
  automation.
\newblock 2021.

\bibitem{CaoNN}
Weidong Cao, Changqing Wang, Weijian Chen, Song Hu, Hua Wang, Lan Yang, and
  Xuan Zhang.
\newblock Fully integrated parity--time-symmetric electronics.
\newblock {\em Nature Nanotechnology}, 17(3):262--268, Mar 2022.

\bibitem{gan}
Petar Veličković, Guillem Cucurull, Arantxa Casanova, Adriana Romero, Pietro
  Liò, and Yoshua Bengio.
\newblock {Graph Attention Networks}.
\newblock In {\em International Conference on Learning Representations}, 2018.

\bibitem{gcn}
Thomas~N. Kipf and Max Welling.
\newblock {Semi-Supervised Classification with Graph Convolutional Networks}.
\newblock In {\em International Conference on Learning Representations}, 2017.

\bibitem{GNN_n0}
Mingjie et~al. Liu.
\newblock {Parasitic-Aware Analog Circuit Sizing with Graph Neural Networks and
  Bayesian Optimization}.
\newblock In {\em 2021 Design, Automation \& Test in Europe Conference
  Exhibition (DATE)}, pages 1372--1377, 2021.

\bibitem{GNN_n1}
Haoxing et~al. Ren.
\newblock {ParaGraph: Layout Parasitics and Device Parameter Prediction Using
  Graph Neural Networks}.
\newblock In {\em Proceedings of the 57th ACM/EDAC/IEEE Design Automation
  Conference}, pages 1--6, 2020.

\bibitem{GNN_distributed}
Guo Zhang, Hao He, and Dina Katabi.
\newblock {Circuit-{GNN}: Graph Neural Networks for Distributed Circuit
  Design}.
\newblock In {\em Proceedings of the 36th International Conference on Machine
  Learning}, pages 7364--7373, 2019.

\bibitem{actor_critic}
Volodymyr Mnih, Adri\`{a}~Puigdom\`{e}nech Badia, Mehdi Mirza, Alex Graves, Tim
  Harley, Timothy~P. Lillicrap, David Silver, and Koray Kavukcuoglu.
\newblock {Asynchronous Methods for Deep Reinforcement Learning}.
\newblock In {\em Proceedings of The 33rd International Conference on Machine
  Learning}, pages 1928--1937, 2016.

\bibitem{ppo}
John Schulman, Filip Wolski, Prafulla Dhariwal, Alec Radford, and Oleg Klimov.
\newblock {Proximal Policy Optimization Algorithms}, 2017.

\bibitem{adam}
Diederik~P. Kingma and Jimmy Ba.
\newblock {Adam: A Method for Stochastic Optimization}.
\newblock {\em CoRR}, abs/1412.6980, 2015.

\bibitem{polar_trans}
Q.~{Diduck} et~al.
\newblock {A 300MHz to 1200MHz Saturated Broadband Amplifier in GaN for 2W
  Applications}.
\newblock In {\em 2016 Texas Symposium on Wireless and Microwave Circuits and
  Systems (WMCS)}, pages 1--4, 2016.

\bibitem{Gallium_nitride}
Stuart Thomas.
\newblock Gallium nitride gets wrapped up.
\newblock {\em Nature Electronics}, 3(12):729--729, Dec 2020.

\bibitem{dgl}
Minjie~Wang et~al.
\newblock Deep graph library: A graph-centric, highly-performant package for
  graph neural networks, 2020.

\end{thebibliography}

\end{document}